\documentclass{mva_style}

\usepackage{graphicx}
\usepackage{times}
\usepackage{epsfig}
\usepackage{graphicx}
\usepackage{amsmath}
\usepackage{amssymb}
\usepackage{multirow}
\usepackage[normalem]{ulem}
\usepackage{comment}
\usepackage{sidecap}
\usepackage{xcolor}
\finalcopy 
\sidecaptionvpos{figure}{t}

\begin{document}
\title{Zero-shot Learning of 3D Point Cloud Objects}



\author{\large Ali Cheraghian, Shafin Rahman and Lars Petersson\\
\large Australian National  University, Data61-CSIRO\\
{\tt\small firstname.lastname@anu.edu.au}}





\maketitle

\section*{\centering Abstract}
\textit{
  Recent deep learning architectures can recognize instances of 3D point cloud objects of previously seen classes quite well. At the same time, current 3D depth camera technology allows generating/segmenting a large amount of 3D point cloud objects from an arbitrary scene, for which there is no previously seen training data. A challenge for a 3D point cloud recognition system is, then, to classify objects from new, unseen, classes. This issue can be resolved by adopting a zero-shot learning (ZSL) approach for 3D data, similar to the 2D image version of the same problem. ZSL attempts to classify unseen objects by comparing semantic information (attribute/word vector) of seen and unseen classes. Here, we adapt several recent 3D point cloud recognition systems to the ZSL setting with some changes to their architectures. To the best of our knowledge, this is the first attempt to classify unseen 3D point cloud objects in the ZSL setting. A standard protocol (which includes the choice of datasets and the seen/unseen split) to evaluate such systems is also proposed. Baseline performances are reported using the new protocol on the investigated models. This investigation throws a new challenge to the 3D point cloud recognition community that may instigate numerous future works.
}

\section{Introduction}
In recent years, a number of methods have been introduced addressing the problem of 3D object classification of point clouds~\cite{Article2,Article24,Article27}. They have achieved outstanding accuracy on available 3D datasets, in most cases reaching a greater performance than $90\%$. This achievement is due to employing deep end-to-end learning on point cloud objects/scenes. Classical attempts of this kind requires numerous pre-processing steps involving voxel representations of 3D models~\cite{Article10}, projecting it to 2D spaces~\cite{Article13}, using pre-trained networks like VGG~\cite{Vgg_arXiv_2014} etc. Those approaches are not only dependent on higher-end hardware but also not extendable to scene understanding/point classification/shape completion. The emergence of end-to-end learning on 3D point cloud data has solved most of the previous problems using one single deep neural network.

\begin{figure}
\centering
\includegraphics[width=.65\linewidth,trim=0cm 0cm 0cm 0cm, clip]{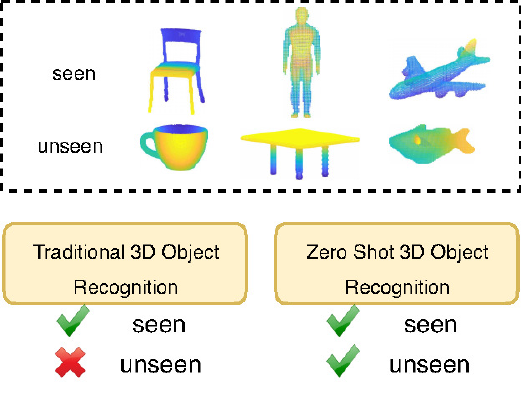}
\caption{\footnotesize Traditional 3D point cloud recognition systems can only classify objects from seen classes. However, adopting a ZSL approach can enable a system to classify objects from classes that are not observed during training.}
\label{fig:overview}
\end{figure}

Nowadays, thanks to the availability of 3D depth cameras, obtaining 3D models of objects and environments are easier than before ~\cite{rs10020328,Izadi_3D_2011}. This has opened up the space of potential applications, and also brings with it new challenges. For example, it is a likely scenario that we have access to, or is capturing, 3D models for which we do not have any labels. Making use of such data is a challenge, and we are in this paper exploring whether a zero-shot learning (ZSL) approach can be utilized (See Figure \ref{fig:overview}). To the best of our knowledge, this is the first time ZSL has been applied to 3D object recognition.
In the 2D image domain, on the contrary, there are many works addressing the ZSL problem in the past few years~\cite{rahman2018unified,Zhang_2017_CVPR,Lee_2018_CVPR,Demirel_2017_ICCV}.  A general ZSL architecture introduces semantic information (like attributes~\cite{Lampert_PAMI_2014} or word vectors~\cite{Mikolov_NIPS_2013,Jeffrey_Glove_2014}) of classes to transfer the knowledge from the seen classes to the unseen classes. During training of ZSL, many methods convert image features to the semantic embedding~\cite{Lampert_PAMI_2014} or a latent space~\cite{Xian_2016_CVPR}, or vice versa~\cite{Zhang_2017_CVPR}. Later, in testing, instances of unseen classes are projected to the same space learned during training, to predict a matching score based on the similarity between the projected embedding and the unseen semantic embedding. With the motivation from the 2D version of the ZSL problem, we use semantic information from classes inside the deep network to classify unseen 3D point cloud objects.

In this paper, we conduct a series of experiments utilising two popular structures traditionally used for feature extraction in 3D point clouds. These are PointNet~\cite{Article1} and EdgeConv~\cite{Article24}. We combine these with two pooling methods, Maxpooling and NetVlad~\cite{Article22}. With the help of these base architectures, we build a new structure that combines point cloud features with word vector semantic features thereby enabling the classification of previously unseen 3D classes.  Theory is here provided as to how to combine semantic word vectors and adapt the previous structure to perform the ZSL task. Moreover, based on our experiments, we demonstrate that our proposed framework for classification of unseen 3D models is useful when the word vector semantic embedding are used during the training stage. Our proposed framework addressing the ZSL task is shown to produce good results on a number of benchmark datasets. In this paper, our main contributions are as follows: 

\begin{itemize}
    \setlength\itemsep{.1em}
   \item We present a new challenge to the ZSL community which aims to classify 3D point cloud objects without having previously observed a single instance of the classes they belong to.
   \item We adapt established 3D point cloud classification methods to perform ZSL which can serve as baseline performance for further research in this direction.
   \item We introduce a new evaluation protocol for ZSL methods on 3D point clouds which consists of a seen and unseen split of data from the datasets ModelNet40~\cite{Article10}, ModelNet10~\cite{Article10}, Mcgill~\cite{Article49} and SHREC2015~\cite{Article48}.
\end{itemize}

\section{Related Work}
\noindent
\textbf{3D point cloud object recognition architecture:} The early methods utilizing deep learning for operating on 3D point clouds used volumetric~\cite{Article10} or multi-view~\cite{Article13} representations in order to work with 3D data. Recently, the trend in this area has shifted to instead using raw point clouds directly~\cite{Article2,Article24,Article27}, without any preprocessing step. These methods do not suffer to the same degree from scalability issues as the volumetric representation does, and they do not make any {\it a priori} assumptions onto which 2D planes, and how many, that the point cloud should be projected on, like the view-based methods do. PointNet~\cite{Article1} was the first work that operated on raw point clouds directly at the input of the network. PointNet used a multi-layer perceptron (mlp)~\cite{Article42} to extract features from point sets, and max-pooling layers to remove the otherwise inherent issue of permutation from the point clouds. Later, many methods~\cite{Article2,Article24,Article27} were proposed to overcome the limitations of PointNet, which does not utilize local features or a more advanced pooling operation than max-pooling. At the time of writing, only the traditional recognition case where all the classes of interest have been seen at training time, have been considered in the case of 3D point cloud data. The current literature does not address the zero-shot version of the 3D recognition problem. In this paper, for the first time, we perform ZSL on 3D point cloud objects. 


\noindent
\textbf{Zero-shot learning on 2D images:} In the image recognition literature, zero-shot learning (ZSL) has made reasonable progress over the past few years~\cite{rahman2018unified,Zhang_2017_CVPR,Lampert_PAMI_2014}. The objective of such learning is to recognize objects from unseen classes not used during training. To do that, semantic information about the class labels in the form of attributes/word vectors are taken advantage of. Image features are usually transferred to the dimension of the semantic vector to obtain a matching score by comparing it with seen/unseen semantic vectors. Some of the notable research directions in this line of investigation include exploring class attribute association~\cite{Demirel_2017_ICCV}, domain adaptation~\cite{Deutsch_2017_CVPR}, the effect of hubness~\cite{Zhang_2017_CVPR}, generalized ZSL~\cite{rahman2018unified}, inductive vs. transductive ZSL ~\cite{Li_2017_CVPR}, multi-label ZSL~\cite{Lee_2018_CVPR} etc. In this paper, we apply zero-shot learning on 3D point cloud objects instead of the traditional 2D image.


\section{Our approach}


\textbf{Problem formulation:} Suppose we define a set of seen $\mathbf{Y}^{s} = \left \{ 1,...,S \right \}$ and a set of unseen $\mathbf{Y}^{u} = \left \{ S+1,...,S+U \right \}$ labels, where $ \mathbf{Y}^{s} \cap  \mathbf{Y}^{u} = \varnothing $, and $S$ and $U$ are the total number of seen and unseen labels respectively. There are associated semantic class embeddings (word vectors) for all samples in both seen and unseen sets, which is defined as $\mathbf{E}^{s}=\left \{ \mathbf{e}_{s}:s\in \mathbf{Y}^{s} \right \}$ and $\mathbf{E}^{u}=\left \{ \mathbf{e}_{u}:u\in \mathbf{Y}^{u} \right \}$ respectively, where $\mathbf{e}_{s},\mathbf{e}_{u}\in \mathbb{R}^{d}$. The number of instances for sets of seen and unseen classes are $n_{s}$ and $n_{u}$ respectively. The matrices $ \mathbf{X}^{s}=\left [{\mathbf{x} _{s}^{1}},...,\mathbf{x} _{s}^{n_{s}} \right ]$ for $s\in \mathbf{Y}^{s}$, and $ \mathbf{X}^{u}=\left [\mathbf{x} _{u}^{1},...,\mathbf{x} _{u}^{n_{u}} \right ]$ for $u\in \mathbf{Y}^{u}$ are point cloud features for the seen and unseen classes respectively, where for both seen and unseen instances $\mathbf{x} =\left \{ \mathbf{p}_{1},...,\mathbf{p}_{n} \right \}$ is an unordered point set, where $\mathbf{p}_{i} \in \mathbb{R}^{F}$. In the simplest setting of $F=3$, each point contains 3D coordinates $x$, $y$, and $z$. To place the problem in a zero-shot setting, it is crucial to mention that, $\mathbf{X}^{u}$, $\mathbf{Y}^{u}$ and $\mathbf{E}^{u}$ are not observed during the training stage. Here, we define the ZSL task addressed in this work: Only point cloud features of seen classes $ \mathbf{X}_{s}$ are used in the training phase. The aim is to assign an unseen class tag $u\in \mathbf{Y}^{u}$ to a given unseen 3D point cloud shape using the related feature vector $\mathbf{x}_{u}$.

\begin{SCfigure} 
\includegraphics[width=.65\linewidth,trim=.4cm 0cm 1.13cm .7cm, clip]{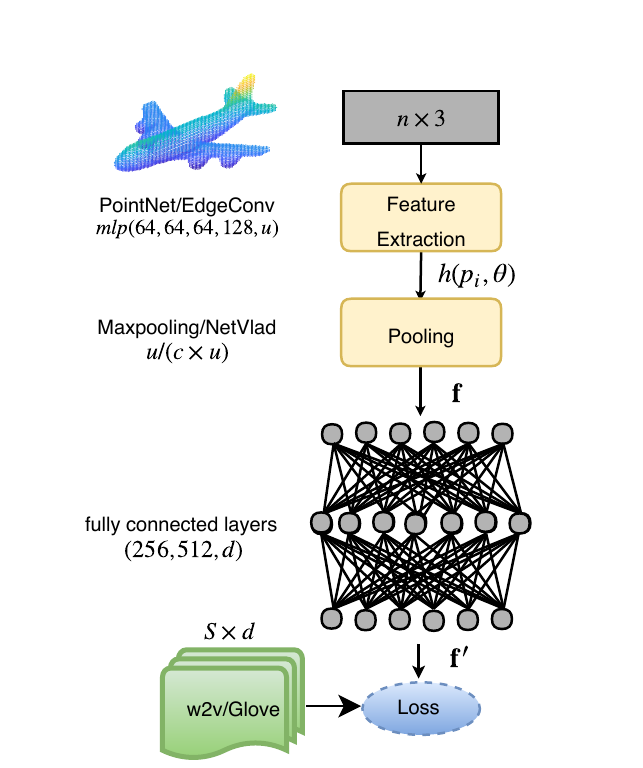}
\hspace{-2em}
\caption{\footnotesize A general framework of the architecture using semantic word vectors. Similar to a traditional 3D point cloud recognition system, it produces a prediction score for every seen object. The inference of unseen classes is made based on those seen predictions.}
\label{fig:architecture}
\vspace{-1cm}
\end{SCfigure}



 \textbf{Training:} Given an unordered point set representing an object from a seen class $\mathbf{x}_{s} =\left \{ \mathbf{p}_{1},...,\mathbf{p}_{n} \right \}$, a set function is defined such that any permutation of the point set becomes irrelevant,
\begin{align*}
f(\mathbf{p}_{1},\mathbf{p}_{2},...,\mathbf{p}_{n}) \approx  g(h(\mathbf{p}_{1},\beta  )),h(\mathbf{p}_{2},\beta  )),...,h(\mathbf{p}_{n},\beta  ))
\end{align*}
where $f$ is the set function, $h$ is the feature extraction function, $g$ is the pooling function with the ability to remove the effects of permutation of points in a set, and $\beta$ represents a set of arguments associated with $\mathbf{p}_i$. The feature extraction function $h(\mathbf{p}_i,\beta )$ extracts a richer representation from the point cloud in a higher dimension. For feature extraction, two different algorithms for feature extraction, $h(\mathbf{p}_{i},\beta )$, are applied. That is, global feature extraction as done in PointNet~\cite{Article1}, and local feature extraction as done in EdgeConv~\cite{Article24}. In PointNet, $h(\mathbf{p}_{i},\beta )=h(\mathbf{p}_{i}) : \mathbb{R}^{d}\rightarrow\mathbb{R}^{d^{\prime}},\beta=\left \{ \emptyset \right \} $, since each point is considered separately, the extracted feature vector contains global information. In EdgeConv~\cite{Article24}, which extracts local features as well as global features, $h(\mathbf{p}_{i},\beta)=h(\mathbf{p}_{i},\mathbf{p}_{j}-\mathbf{p}_{i}) : \mathbb{R}^{d}\times \mathbb{R}^{d}\rightarrow\mathbb{R}^{d^{\prime}},\beta =\left \{ \mathbf{p}_{j}-\mathbf{p}_{i} \right \} $. In this case, point sets are represented by a dynamic graph and edge features based on $k$-nearest neighbors are calculated. Since point sets are inherently unordered, a function which is invariant to permutation is necessary to pool point features into a feature vector. Here, the Maxpooling operation above, $g$, is capable of removing the permutation from point clouds. Also, instead of Maxpooling,~\cite{Article37} used NetVlad~\cite{Article22} as a pooling mechanism to remove permutation from point cloud features.


Finally, via a collection of $h(\mathbf{p}_i,\beta )$, corresponding values of $f$ can be computed to form a vector $\mathbf{f}\in \mathbb R^{m}$ for Maxpooling pooling, and $\mathbf{f}\in \mathbb R^{m \times c}$ for NetVlad pooling, where $c$ is the number of centers in NetVlad, which represents a feature vector of an input point set. The obtained feature vector removes permutation from the point cloud. In the next step, a few fully-connected layers are applied to the feature vector $\mathbf{f}$ in order to transform the features into a more discriminative representation, $\mathbf{{f}'}=\phi(\mathbf{f})$, where $\phi:\mathbb R^{m}\rightarrow \mathbb R^{d}$ represents three fc layers with a nonlinear relu activation. This function maps the point cloud feature vector $\mathbf{f}$ to a semantic word vector embedding space by calculating $\mathbf{{f}'}$. Then, semantic embedding vectors $\mathbf{E}^{s}$, from all seen classes, are inserted into the network by multiplying with the feature vector $\mathbf{f{}'}$ from the fully connected layers, $\mathbf{f{}''}=\mathbf{E}^{s}.\mathbf{f{}'}$, where $\mathbf{E}^{s} \in \mathbb{R}^{ S \times d}$, and $\mathbf{f{}''} \in \mathbb{R}^{ S }$. Finally, the following objective function is minimized in order to train the proposed method shown in Figure \ref{fig:architecture}:
\vspace{-.4cm}
\begin{align*}
L = -\sum_{i=1}^{S}Y_{i}^{s}log(y_{i}))\quad \textrm{where,} \, y_{i}=\frac{e^{(E^{s}.f{}'_{i})}}{\sum_{j=1}^{S}e^{(E^{s}.f{}'_{j})}}
\end{align*}

\noindent
\textbf{Multiple semantic space fusion:} In Figure \ref{fig:architecture}, only one semantic representation, w2v or glove, is considered during the training stage. However, it is also possible to fuse both semantic representations and consider them a unit semantic vector by concatenating the two different semantic embedding space representations. Therefore, the new semantic vector is: $E^{s} = concat(E^{s_{w2v}},E^{s_{glove}})$, where $\mathbf{E}^{s_{w2v}}=\left \{ \mathbf{e}_{s_{w2v}}:s\in \mathbf{Y}^{s} \right \}$,  $\mathbf{E}^{s_{glove}}=\left \{ \mathbf{e}_{s_{glove}}:s\in \mathbf{Y}^{s} \right \}$, and ``$concat$" denotes the concatenation operator.


\textbf{Inference:} During the training stage, a classifier $p_{s}$ is trained on seen instances $\mathbf {X}^{s}$ such that it can estimate the probability of a 3D point cloud $\mathbf{x}$ belonging to a certain seen class label $s \in \mathbf{Y}^{s}$, denoted $p_{s}(s|\mathbf{x}_{s})$, where $\sum_{s=1}^{S}p_{s}(s|\mathbf{x}_{s})=1$. Then, given $p_{s}$, we apply a method similar to~\cite{norouzi_arXiv_2013}, to transfer those probabilities obtained from training samples to the testing labels. 
Suppose $\hat{s}(\mathbf{x},t)$ is the $t^{th}$ most likely label for a point cloud $\mathbf{x}$:
\vspace{-.4cm}

\begin{align}
\hat{s}(\mathbf{x},t)=\underset{s\in Y^{s}}{\arg \max} \, p_{s}(s|\mathbf{x}_{s}) \label{eq:prediction}
\end{align}
\vspace{-.3cm}

So, $p_{s}(\hat{s}(\mathbf{x},t)|\mathbf{x})$ is the $t^{th}$ largest value in $\left \{  p_{s}(s|\mathbf{x});s \in Y^{s}\right \}$. Based on the top $T$ predictions of $p_{s}$, the proposed model calculates an embedding semantic feature vector $z(\mathbf{x})$ for an input $\mathbf{x}$, 
\vspace{-.4cm}

\begin{align}
z(\mathbf{x}) = \frac{1}{K}\sum_{t=1}^{T}p_{s}(\hat{s}(\mathbf{x},t)|\mathbf{x}).e_{st}
\label{eq:topseen}
\end{align}
\vspace{-.4cm}

where $T$ is a parameter in order to control the maximum number of embedding vectors that contributes in the inference stage, and $K=\sqrt{\sum_{t=1}^{T}( p_{s}(\hat{s}(\mathbf{x},t)|\mathbf{x}) )^2}$ is a normalization factor. The significance of using a subset of seen prediction values is to describe unseen classes via only closely related seen classes. Based on the predicted embedding of $\mathbf{x}$ in the semantic space, $z(\mathbf{x})$, zero-shot classification is applied to find the class which is nearest to the obtained embedding vectors. The top prediction for a point cloud $\textbf{x}$ from the test set, denoted $\hat{y}(\mathbf{x},1)$, is defined by using the cosine similarity to rank the embedding vectors, 
\vspace{-.4cm}

\begin{align}
\hat{y}(\mathbf{x},1)=\underset{u\in Y^{u}}{\arg \max} \, cos(z(\mathbf{x}),E^{u}) 
\end{align}
\vspace{-.4cm}

\begin{figure}[!t]
\centering
\includegraphics[width=1\linewidth,trim=1.2cm 2.5cm 0 0, clip]{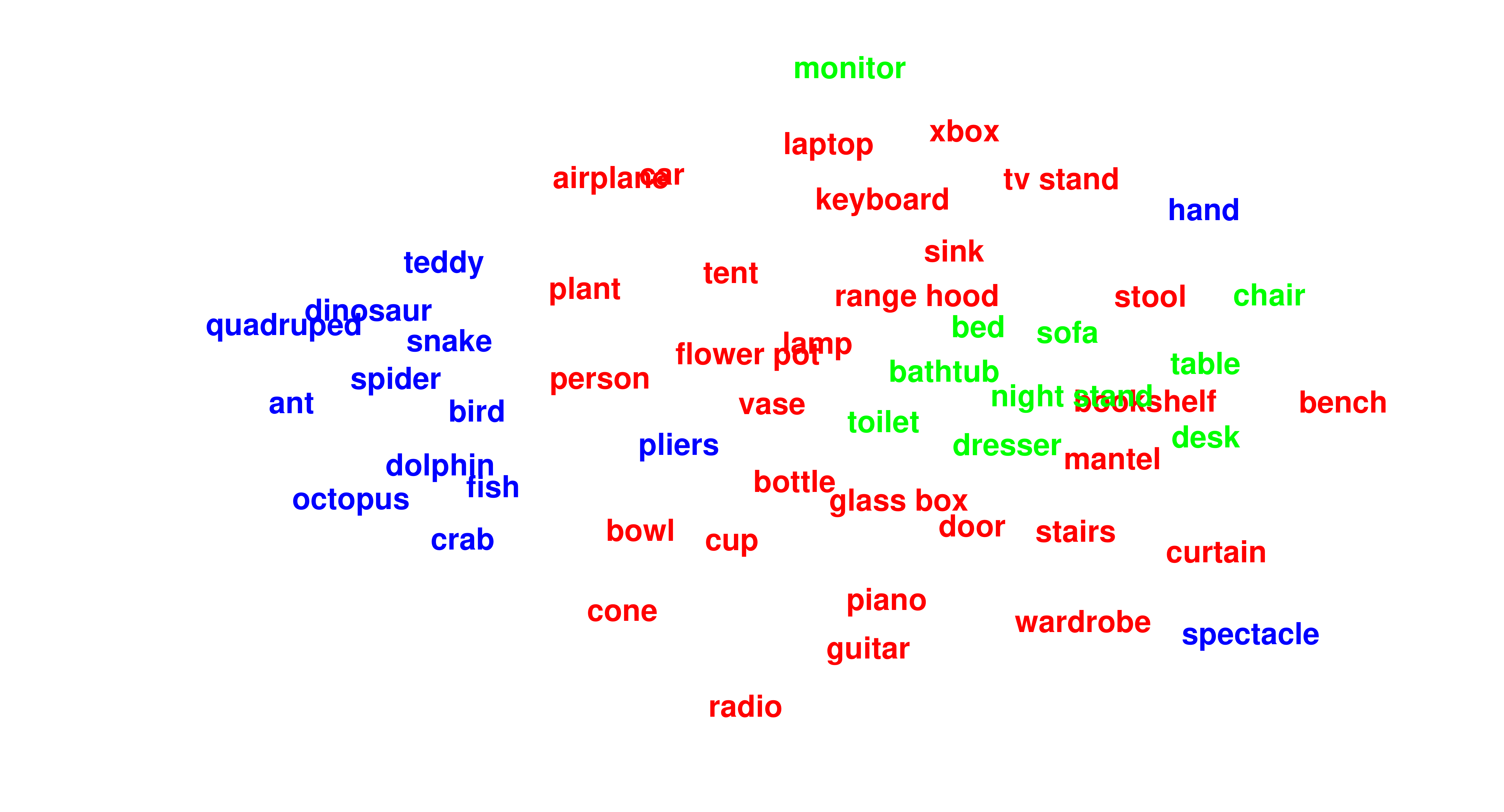}
\caption{\footnotesize 2D tSNE~\cite{tSNE_van2014} visualization of word2vec vectors \cite{Mikolov_arXiv_2013}.  Red, green and blue texts represent seen {\color{red}ModelNet40 \cite{Article10}}, unseen {\color{green}ModelNet10 \cite{Article10}} and unseen {\color{blue}McGill \cite{Article49}} classes respectively.}

\label{fig:w2vglo}
\vspace{-2em}
\end{figure}

\textbf{An alternative approach to adapt ZSL:} In this paper, we modified the PointNet~\cite{Article1} and EdgeConv~\cite{Article24} architectures such that they can consider word vectors during training. This modification helps aligning the point cloud object features to their corresponding semantic space. The output of this architecture still provides a prediction value, $p_{s}(s|\mathbf{x}_{s})$ for each seen class which we use during inference in Eq. \ref{eq:prediction}. However, as the original PointNet~\cite{Article1} and EdgeConv~\cite{Article24} architectures can also provide such seen prediction values, one can consider the original framework instead of our modified architecture. Then, the inference follows the same process as mentioned above. We name this alternative process of prediction the {\it basic} approach. In the experiments, we find that as this basic approach does not consider word vectors during training, it does not perform well predicting unseen classes (See Table \ref{Table:Accuracy}). However, being free from the noise inside the word vectors, it performs very well recognizing objects of seen classes (See Table \ref{Table:SeenClasses}).

\begin{table*}[!ht]
\centering
\caption{Overall Top-1 accuracy on unseen classes. Random accuracy is calculated by $\frac{100}{\# \, of \, unseen \, \, classes}$}
\scalebox{.9}{
\begin{tabular}{|c|c|c|c|c|c|c|c|c|c|c|c|c|}
\hline
\multirow{2}{*}{Method} & \multicolumn{4}{c|}{ModelNet10} & \multicolumn{4}{c|}{McGill} & \multicolumn{4}{c|}{SHREC2015} \\ \cline{2-13} 
 & \multicolumn{1}{c|}{basic} & \multicolumn{1}{c|}{w2v} & \multicolumn{1}{c|}{glove} & \multicolumn{1}{c|}{conc} & \multicolumn{1}{c|}{basic} & \multicolumn{1}{c|}{w2v} & \multicolumn{1}{c|}{glove} & \multicolumn{1}{c|}{conc} & \multicolumn{1}{c|}{basic} & \multicolumn{1}{c|}{w2v} & \multicolumn{1}{c|}{glove} & \multicolumn{1}{c|}{conc} \\ \hline
PointNet\cite{Article1}+ Maxpooling & 23.1 & 27.0 & 14.8 & 19.4 & 14.1 & 9.8 & 7.2 & 11.6 & 3.1 & 4.1 & 3.6 & 4.2 \\ \hline
PointNet\cite{Article1} + NetVlad\cite{Article22} & \multicolumn{1}{c|}{9.1} & \textbf{28.0} & 20.9 & 20.5 & 4.5 & 10.7 & 10.7 & \textbf{16.1} & 3.1 & 5.2 & 4.2 & \textbf{6.8} \\ \hline
EdgeConv\cite{Article24} + Maxpooling & 21.2 & 24.4 & 14.1 & 16.9 & 12.3 & 8.9 & 9.8 & 8.0 & 3.3  & 6.2 & 4.7 & 5.2 \\ \hline
EdgeConv\cite{Article24} + NetVlad\cite{Article22} & 9.4 & 14.6 & 12.7 & 18.5 & 6.5 & 9.6 & 7.8 & 8.1 & 3.0 & 5.2 & 4.1 & 4.1 \\ \hline \hline
Random & \multicolumn{4}{c|}{10.0} & \multicolumn{4}{c|}{7.1} & \multicolumn{4}{c|}{3.3} \\ \hline
\end{tabular}}
\label{Table:Accuracy}
\end{table*}

\begin{table}[!ht]
\centering
\caption{Overall Top-1 accuracy on seen classes.}
\vspace{-.8em}
\scalebox{.85}{
\begin{tabular}{|c|c|c|c|c|}
\hline
 Method& \multicolumn{1}{c|}{basic} & \multicolumn{1}{c|}{basic} & \multicolumn{1}{c|}{w2v} & \multicolumn{1}{c|}{glove} \\
 & (40) & (30) & (10) & (10) \\ \hline 
PointNet\cite{Article1}+Maxpooling & 89.2 & 89.5 & 87.4 & 87.6 \\ \hline
PointNet\cite{Article1}+NetVlad\cite{Article22} & 87.1 & 87.7 & 81.2 & 81.2  \\ \hline
EdgeConv\cite{Article24}+Maxpooling & 92.2 & 92.3  &  90.7 & 89.5 \\ \hline
EdgeConv\cite{Article24}+NetVlad\cite{Article22} & 91.2 & 91.4 & 86.0 & 83.8 \\ \hline
\end{tabular}}
\label{Table:SeenClasses}
\end{table}

\section{Experiment}

\subsection{Setup}
\noindent
\textbf{Dataset:} We perform our experiments on three well-known 3D datasets, ModelNet40/Modelnet10~\cite{Article10}, McGill~\cite{Article49}, and SHREC2015~\cite{Article48}. The ModelNet40 contains $12,311$ CAD models from 40 different classes, and Modelnet10, which is a subset of ModelNet40, consists of $4899$ CAD models from 10 different classes. The Modelnet40 contains $9,843$ training samples and $2,468$ testing samples, and ModelNet10 includes $3991$ training samples and $908$ testing samples. Moreover, the McGill dataset consists of 456 CAD models of 19 different objects. The SHREC2015 dataset consists of 1200 3D watertight triangle meshes from 50 different classes, where each class contains 24 objects with distinct postures. Each shape in the dataset has approximately 10,000 vertices. Note that, we randomly sample all models to $1,024$ points from the mesh faces and normalize them to a unit sphere. We use the (x, y, z) coordinates of the sampled points for all the experiments in this work.



\noindent
\textbf{Seen/Unseen split:} We are proposing a suitable and standard split of seen and unseen classes similar to the 2D version of the corresponding ZSL problem. In the 2D case, experiments using established methods of ZSL are performed on a single, well defined, split of seen/unseen classes from datasets like Animals with Attributes (AwA)~\cite{AwA_2009}, and Caltech-UCSD Birds (CUB)~\cite{CUB_2011}. The idea of using a single split is better than using several random splits as it establishes a common testbed for evaluating other methods. In a real life setting, unseen 3D point cloud objects are likely obtained from an advanced 3D-depth camera or laser scanner. Here, however, we attempt to simulate that scenario by using 3D point cloud objects from a different dataset not included during training. Therefore, we propose to use 30 classes from ModelNet40 as seen and other, disjoint classes, from ModelNet10, McGill and SHREC2015 as unseen. In ModelNet10, which has 10 classes, only testing samples based on the splitting protocol mentioned above, are considered as the unseen set. In McGill~\cite{Article49}, 5 classes, which also appear in the seen set, are removed, and the remaining 14 categories are selected. Since we follow the protocol introduced by~\cite{Article29}, we only consider their test set, the last third of the instances, as the unseen instances to enable a fair comparison. Finally, in the SHREC2015~\cite{Article48} dataset, those classes that were similar to the seen classes, and those classes for which there were no semantic word vector available, were removed. The remaining 30 of the original 50 classes were chosen as the unseen instances. Moreover, in line with the protocol used in~\cite{Article28}, 25\% of instances were chosen randomly as unseen samples. 


\noindent
\textbf{Semantic features:} 
We work with unsupervised word vectors obtained from an unannotated text corpus as semantic information. We used  $\ell_2$ normalized, 300 dimensional word2vec~\cite{Mikolov_NIPS_2013} and Glove~\cite{Jeffrey_Glove_2014} word vectors. The 2D tSNE visualization~\cite{tSNE_van2014} of those vectors is illustrated in Figure~\ref{fig:w2vglo}.

\noindent
\textbf{Evaluation:} The recognition performance is here measured by top-$1$ accuracy. It means the class with the highest predicted probability must match the ground truth class to be considered ``correct".
\noindent
\textbf{Implementation details:} We use the Adam optimizer and a batch size of 16.  We also use Relu and Batch normalization(BN)~\cite{Article45} for each layer. We implemented the architecture using TensorFlow and executed on an NVIDIA GTX1080Ti. For PointNet, we employed five shared mlp layers (64,64,64,128,1024). Also, for EdgeConv, which consists of two feature extraction blocks, the first block used three shared mlp layers (64,64,64) and the second block consisted of a shared mlp of size (128). Then, a shared mlp of size 1024 was used to concatenate all features together. Finally, for aggregation, we also used either max pooling or NetVlad. In the case of max pooling, we got a feature vector of size 1024. In the case of NetVlad, the number of cluster centers was set to 128 and the last mlp layer was set to size 128, from which we got a feature vector of size $128\times128$. 



\begin{SCfigure}
\includegraphics[width=.75\linewidth,trim=.5cm 0cm 1.4cm .4cm, clip]{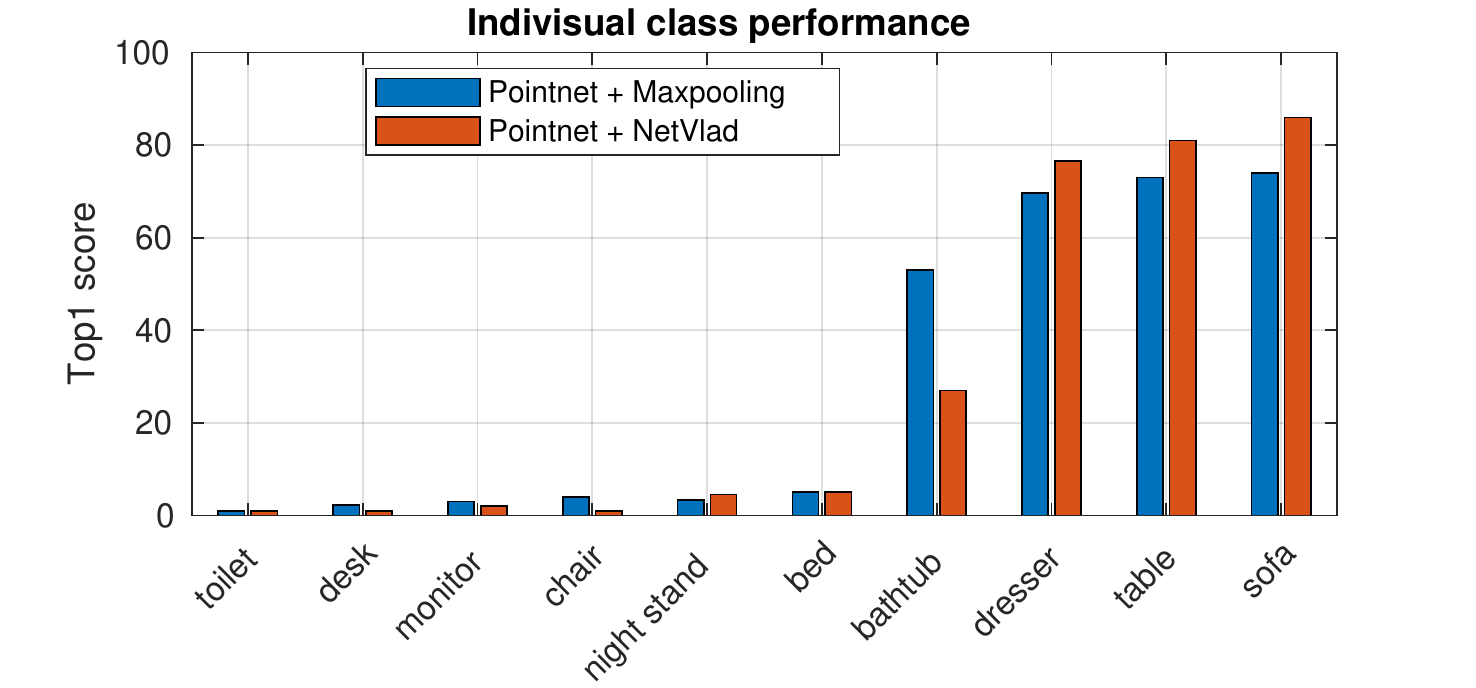}
\hspace{-2em}
\caption{\footnotesize Per-class recognition rate of unseen ModelNet10 classes.}
\label{fig:IndivitualClassPerformance}
\vspace{-6em}
\end{SCfigure}

\subsection{Recognition results}

\textbf{Unseen recognition:} In this subsection, the top-1 accuracy performance of four different structures based on the combinations of feature extraction modules PointNet and EdgeConv, and the pooling layers Maxpooling and NetVlad, are evaluated. Moreover, for each structure, four different experiments, {\it basic, w2v, glove,} and {\it conc}atenation of w2v and glove, are conducted. It is important to mention that in the basic experiments, non semantic word vector embeddings are used during the training stage. As shown in Table~\ref{Table:Accuracy}, the winning architecture for point cloud zero-shot learning is the combination of PointNet and NetVlad, which achieves an accuracy of 28.9\%, 16.1\%, and 6.8\% for the ModelNet10, McGill, and SHREC2015 datasets respectively. Based on the obtained results, we can make the following observations: 1) PointNet, in comparison to EdgeConv, has the advantage and performs better on the ZSL task, which can be due to the fact that the vanilla PointNet version has a simpler structure and fewer parameters to be learned. However, the EdgeConv module is better suited to the non-ZSL, point cloud recognition task than PointNet. 2) NetVlad works better than Maxpooling. Although Maxpooling is robust to permutation, it is a greedy operation which removes much useful information. In comparison to the Maxpooling layer, NetVlad keeps more features and, similarly, it ignores permutation in point cloud data. 3) Zero-shot 3D object recognition benefits from using a semantic word vector embedding in all investigated structures in comparison to the basic model. 4) Concatenation of w2v and glove helped to improve the performance of the proposed method on the McGill and SHREC2015 datasets.

\textbf{Per-class results:} In Figure~\ref{fig:IndivitualClassPerformance}, we report individual class performance of unseen ModelNet10 classes using PointNet + Maxpooling/NetVlad. Our architecture performs better on ModelNet10 classes (e.g., sofa, table, dresser, and bathtub) for which there are sufficiently similar classes in the seen set. However, one can find that the architecture cannot classify some unseen classes at all (for example toilet, crab, dinosaur etc). This is due to the hubness problem in high dimensional space Zhang \textit{et al.}~\cite{Zhang_2017_CVPR}.

\textbf{Seen recognition:} To investigate the effect of using semantic feature vectors during the training stage, in the case of the traditional non-ZSL task, we also evaluated seen class performance (see Table~\ref{Table:SeenClasses}). To this end, the test set of seen classes of the ModelNet40 splitting protocol are considered. Also, we tested basic models with 30 and 40 seen classes. However, the performance of the basic model is slightly better than the models which used semantic word vectors during training. Reasonably, being obtained in an unsupervised way, word vectors insert noise in the training which reduces the seen recognition rate but helps the unseen recognition rate.


\begin{SCfigure}
\includegraphics[width=.7\linewidth,trim=1.4cm 0cm 1.1cm .4cm, clip]{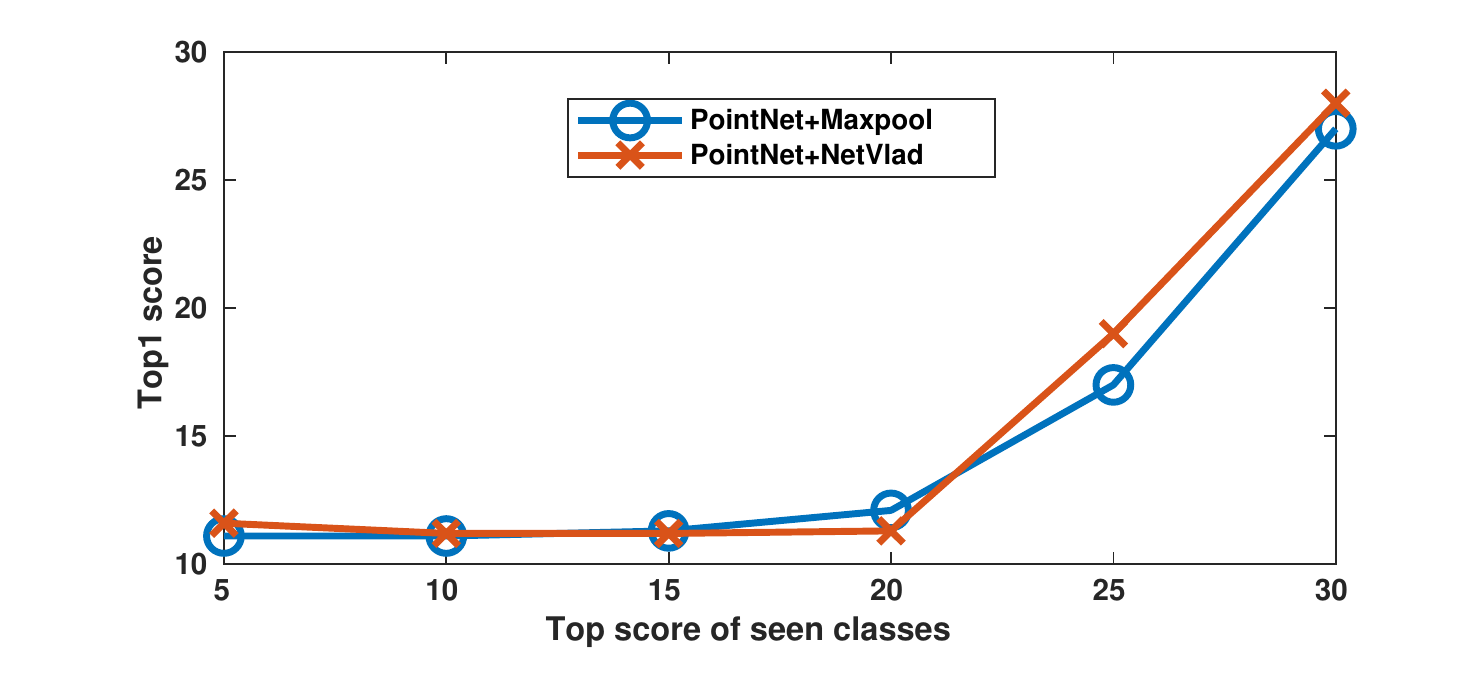}
\hspace{-2em}
\caption{\footnotesize The effect of varying $T$, the number of embedding vectors in Eq. \ref{eq:topseen}.}
\label{fig:TopScore}
\vspace{-6em}
\end{SCfigure}





\textbf{Parameter Dependency:} Eq. \ref{eq:topseen} allows us to vary the number of embedding vectors from the seen classes that contribute in the inference stage. In Figure \ref{fig:TopScore}, we see the result when varying this number from 5 to 30. Note that, in Fig. \ref{fig:w2vglo}, the ModelNet10 vectors reside within the distribution of the ModelNet40 vectors . Hence, most seen ModelNet40 vectors contribute to describing unseen ModelNet10 objects. Therefore, ModelNet10 performs better with a large number of seen vectors.

\vspace{-0.2cm}
\section{Conclusion}
Traditional recognition systems have achieved superior performance on 3D point cloud objects. However, due to the advancement of 3D depth camera technology, obtaining 3D point cloud representations of scenes has become much more accessible than before. Hence, we will more than likely encounter many unseen objects to which our traditional 3D point cloud recognition gets no training. Therefore, it is time for 3D point cloud recognition systems to adapt zero-shot settings, aiming to recognize those unseen objects. To this end, this paper proposed a new challenge and a useful evaluation testbed/protocol for pushing forward with this new line of investigation. We also modified some established 3D point cloud recognition systems to work in the zero-shot setting in order to report a set of baseline performance results with respect to this problem. Overall, we believe that this research has the potential to motivate numerous further works to create a more robust 3D point cloud recognition system.

{
\bibliographystyle{ieee}
\bibliography{egbib}
}







\end{document}